\documentclass[sn-mathphys-num]{sn-jnl}

\usepackage[table]{xcolor}
\usepackage{graphicx}%
\usepackage{multirow}%
\usepackage{amsmath,amssymb,amsfonts}%
\usepackage{amsthm}%
\usepackage{mathrsfs}%
\usepackage[title]{appendix}%
\usepackage{textcomp}%
\usepackage{manyfoot}%
\usepackage{booktabs}%
\usepackage{algorithm}%
\usepackage{algorithmicx}%
\usepackage{algpseudocode}%
\usepackage{listings}%
\usepackage{footnote}
\usepackage{amsmath}
\usepackage{amssymb}
\usepackage{amsfonts}
\usepackage{graphicx}
\usepackage{textcomp}
\usepackage{amsmath}
\usepackage{algorithm}
\usepackage{array}%
\usepackage{siunitx}%
\usepackage{multirow}

\usepackage{tikz}
\usepackage{nomencl}
\usepackage{amsmath,amssymb}
\usepackage{natbib}
\usepackage{graphicx}%
\usepackage{caption}%
\makenomenclature
\renewcommand{\nomname}{Nomenclature}
\usepackage{makecell}
\definecolor{highlight}{rgb}{0.96, 0.76, 0.76} %

\begin{document}
\title{Detecting Zero-Day Attacks in Digital Substations via In-Context Learning}


\author*[1]{\fnm{Faizan} \sur{Manzoor}}\email{mfaizan@vt.edu}

\author[1]{\fnm{Vanshaj} \sur{Khattar}}\email{vanshajk@vt.edu}

\author[1]{\fnm{Akila} \sur{Herath}}\email{akilaasansana@vt.edu}

\author[2]{\fnm{Clifton} \sur{Black}}\email{crblack@southernco.com}

\author[3]{\fnm{Matthew C} \sur{Nielsen}}\email{nielsema@ge.com}

\author[4]{\fnm{Junho} \sur{Hong}}\email{jhwr@umich.edu}

\author[1]{\fnm{Chen-Ching} \sur{Liu}}\email{ccliu@vt.edu}

\author*[1]{\fnm{Ming} \sur{Jin}}\email{jinming@vt.edu}

\affil[1]{\orgdiv{Electrical and Computer Engineering Department}, \orgname{Virginia Tech}}

\affil[2]{\orgname{Southern Company}}

\affil[3]{\orgname{GE Global Research}}

\affil[4]{\orgdiv{Electrical and Computer Engineering Department}, \orgname{University of Michigan-Dearborn}}

\abstract{The occurrences of cyber attacks on the power grids have been increasing every year, with novel attack techniques emerging every year. In this paper, we address the critical challenge of detecting novel/zero-day attacks in digital substations that employ the IEC-61850 communication protocol. While many heuristic and machine learning (ML)-based methods have been proposed for attack detection in IEC-61850 digital substations, generalization to novel or zero-day attacks remains challenging. We propose an approach that leverages the in-context learning (ICL) capability of the transformer architecture, the fundamental building block of large language models. The ICL approach enables the model to detect zero-day attacks and learn from a few examples of that attack without explicit retraining. Our experiments on the IEC-61850 dataset demonstrate that the proposed method achieves more than $85\%$ detection accuracy on zero-day attacks while the existing state-of-the-art baselines fail. This work paves the way for building more secure and resilient digital substations of the future.}


\renewcommand{\nomname}{Nomenclature} 

\renewcommand{\nomgroup}[1]{%
  \ifthenelse{\equal{#1}{A}}{\item[\textbf{Model Parameters and Dimensions:}]}{%
  \ifthenelse{\equal{#1}{B}}{\item[\textbf{Data Structure and Input Representations:}]}{%
  \ifthenelse{\equal{#1}{C}}{\item[\textbf{Individual Classifier Outputs:}]}{%
  \ifthenelse{\equal{#1}{D}}{\item[\textbf{Combined Classifier Representations:}]}{%
  \ifthenelse{\equal{#1}{E}}{\item[\textbf{Mixed Label/Distribution Representations:}]}{\ifthenelse{\equal{#1}{F}}{\item[\textbf{Model Deployment:}]}{}}}}}}}%

\maketitle



\section{Introduction}\label{sec1}
The modern-day power grids continuously face the possibility of being disrupted by cyber attacks. This is evident from the recent increase in the cases of cyber attacks on power grids worldwide. For example, in 2015, a coordinated cyber-attack was responsible for the mass-scale power outages in Ukraine \cite{assante2016confirmation}. In 2016, another cyber-attack in Ukraine also led to a mass power outage and affected the SCADA system at the transmission level \cite{case2016analysis}.

Intrusion Detection Systems (IDSs) play an important role in detecting potential cyber-attacks on substations so that timely action can be taken. However, most of the existing IDS methods, heuristic or machine learning-based, struggle when they encounter a novel or an unseen attack. We refer to these kinds of attacks as \emph{zero-day attacks} as they present themselves for the first time. A recent study showed that millions of new cyber-attacks were detected annually worldwide from 2015 to 2020 \cite{GlobalReport}.  
This highlights the importance of developing IDSs that can effectively detect zero-day attacks. As a consequence, cybersecurity of digital substations against zero-day attacks has recently been a focus for many researchers \cite{duman2017measuring}.

In this paper, we specifically consider IEC-61850 digital substations \cite{lozano2023digital,10688802}. Although IDS techniques are well-explored for traditional TCP/IP-based substation communication networks \cite{stefanov2012cyber}, the specific requirements and unique communication protocols of IEC–61850 substations have not been adequately addressed. 

The IEC–61850 communication protocol is commonly used in digital substations to communicate between Intelligent Electronic Devices (IEDs) and Merging Units (MUs). It allows for efficient connectivity and control in digital substations \cite{international2013communication}, however, there are numerous vulnerabilities that an attacker can exploit to disrupt the operation of these digital substations \cite{otuoze2018smart}.  
Many existing IDS methods, whether heuristic-based \cite{hong2014integrated} or machine learning (ML)-based \cite{sahani2023machine, park2024machinelearningbasedcyber}, are designed for specific attacks, which can limit their ability to generalize to zero-day attacks.

Heuristic-based methods use a predefined set of rules to determine if the incoming data packet is a specific attack or not. In \cite{hong2014detection}, the authors use timestamp and sequence numbers to detect a replay attack. In \cite{rajkumar2020cyber}, authors investigated the injected spoofing attack on the IEC-61850-based standard. However, the proposed model was specifically designed for spoofing attacks and may have limited applicability to other types of attacks. The work in \cite{yang2016intrusion} develops an IDS for the IEC-61850 protocol, but they only consider information carried within sampled value messages, which restricts its application to other message-sharing protocols, such as GOOSE.

In \cite{el2019iec}, the authors use a neural network-based approach to detect spoofed packets. Another work improved the previous approach using the decision trees and random forests \cite{ustun2021artificial}. However, many of these existing ML-based methods focus on specific attack cases and do not generalize to novel attacks. 

Therefore, we propose a generalizable IDS framework for the IEC-61850 communication protocol that can detect zero-day attacks on digital substations. Our method leverages the ``in-context learning" (ICL) ability of transformer architectures \cite{vaswani2017attention}, which are a basic building block of large language models (LLMs) \cite{brown2020language}. 

In-context learning is the ability to generalize rapidly from a few examples of a new task that have not been previously seen, without any updates to the model, a key characteristic of many large language models (LLMs) \cite{brown2020language,min2022metaicllearninglearncontext}. For example, consider the following context examples of network packets provided to an LLM: $Packet1 = Normal; Packet2 = Normal; Packet3 = Attack$. Then, if the query sample to the LLM is $Packet4$, which shares similar characteristics with $Packet3$, the LLM may output ``Attack" as it is able to understand through the in-context examples that packets with certain features are classified as attacks. This ability of LLMs to understand the context and adapt their outputs accordingly without any additional training motivates the following question:
 
 \textbf{``Can this in-context learning ability be leveraged for zero-day attack detection?''} 

 In this paper, we answer the above question in the affirmative and show that the transformer architecture could use the incoming data packets in a digital substation as in-context examples to infer if the next incoming packet is an attack or not. 
 
 We begin by splitting the dataset into in-distribution (ID) (known dataset) and out-of-distribution (OOD) (zero-day attacks) subsets, using the ID portion for training and reserving the OOD data exclusively for testing. To expand the variety of potential attack classes within the ID set, we employ the multi-mixing \cite{manzoor2024zero} strategy to generate additional attack classes.

Next, we train neural network–based weak classifiers on the augmented ID dataset. These weak classifiers provide weak-labels—either as hard labels (Lab) or probability distributions (Dist)—that our transformer architectures use during both training and inference. Specifically, the models process historical packet data and their Lab/Dist (in-context data)  alongside the latest packet (the query), leveraging ICL to predict whether the query packet is anomalous.

We instantiate two main transformer variants:
\begin{enumerate}
    \item \textbf{Simple Transformer (TF)}, which conditions on hard labels, and
\item \textbf{Distributional Transformer (DTF)}, which conditions on label distributions.
\end{enumerate}

Both variants are trained using standard cross-entropy loss, but with two distinct strategies for handling in-context Lab/Dist: (1) using only the weak classifier, or (2) mixing the weak classifier with ground-truth Lab/Dist for the entire in-context data. This yields four specific training strategies:

\begin{enumerate}
    \item \textbf{Weak Classifier Trained Simple Transformer (WCTF)}: TF trained with in-context hard labels generated from weak classifiers.
\item \textbf{Mixed Trained Simple Transformer (MTF)}: TF trained with in-context hard labels generated from weak classifiers and ground truth.
\item \textbf{Weak Classifier Trained Distributional Transformer (WCDTF)}: DTF trained with in-context label distribution generated from the weak classifiers.
\item \textbf{Mixed Trained Distributional Transformer (MDTF)}: DTF trained with in-context label distribution generated from weak classifiers and ground truth.
\end{enumerate}

In real-world scenarios, ground-truth labels are not available; hence, during testing, Lab/Dist for in-context data are generated solely from weak classifiers. Despite this limitation, our framework effectively identifies zero-day attacks by capitalizing on the ICL capabilities of modern transformer architectures.

Our approach for zero-day attack detection also shares similarities with several advanced ML paradigms, including transfer learning \cite{weiss2016survey}, OOD detection \cite{yang2021generalized, abdeen2024defensejointpoisonevasion}, meta-learning \cite{khattar2022cmdp, 10738096,pmlr-v211-sel23a}, and multi-task learning \cite{zhang2018overview}. Each of these aims to enhance the generalization capability of ML models and enable them to adapt to unseen data. However, our work specifically exploits the ICL capability of transformer architectures, allowing them to adapt to new tasks based solely on input context without further retraining. Closest to our approach is our previous work \cite{manzoor2024zero}, where we initially leveraged ICL to detect zero-day attacks. In this paper, we extend that effort by introducing a distributional transformer architecture that accounts for the distribution of labels associated with each in-context data packet (refer Figure~\ref{fig:DTFvsTF}), rather than relying on hard labels to identify the most probable attack scenario. In addition, we perform additional ablation studies to investigate the impact of various design choices on the effectiveness of our proposed framework.


\begin{figure*}[!ht]
    \centering 
    \includegraphics[width=1\textwidth]{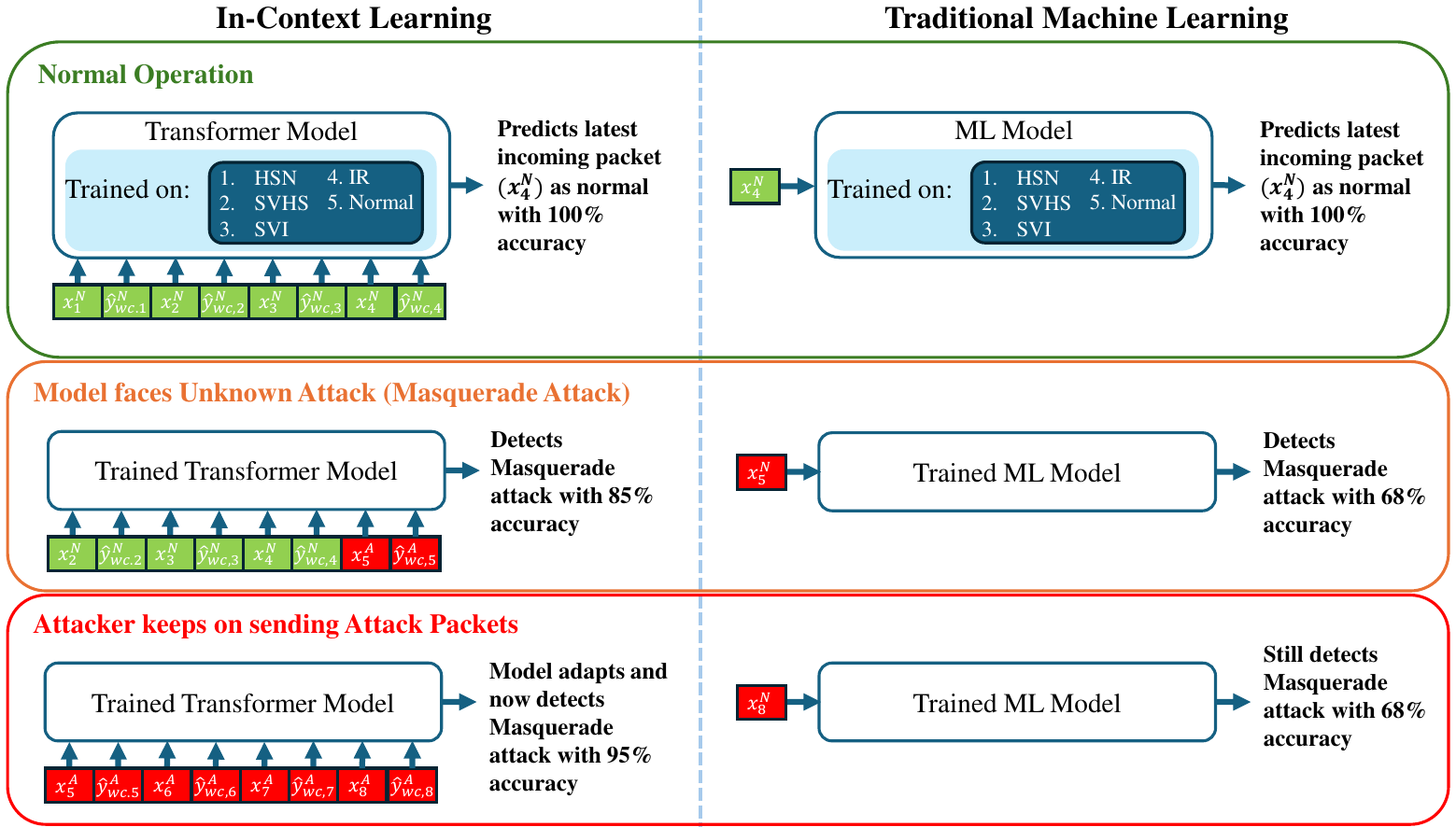} 
    \caption{Traditional ML models, such as Random Forest, Decision Trees, and Convolutional Neural Networks trained on a specific dataset (GOOSE High Status Number (HSN), Inverse Replay (IR), SV High Status Number (HSN), SV Injection (SVI)), often fail to identify novel attacks during the deployment phase. To adapt to the new threats, they require retraining with datasets that include these novel attacks. 
    In contrast, the proposed ICL-based approach can detect novel attacks even if they were not included in the training dataset. ICL allows it to use the in-context data and weak labels $(\hat{y}_{wc,i})$ to better generalize and recognize new attacks without the need for retraining or parameter updates.}
    \label{fig:MLvsTF_comparison}
    \vspace{-0.4cm}

\end{figure*}

\begin{figure}[p]
    \centering
    \includegraphics[width=\columnwidth]{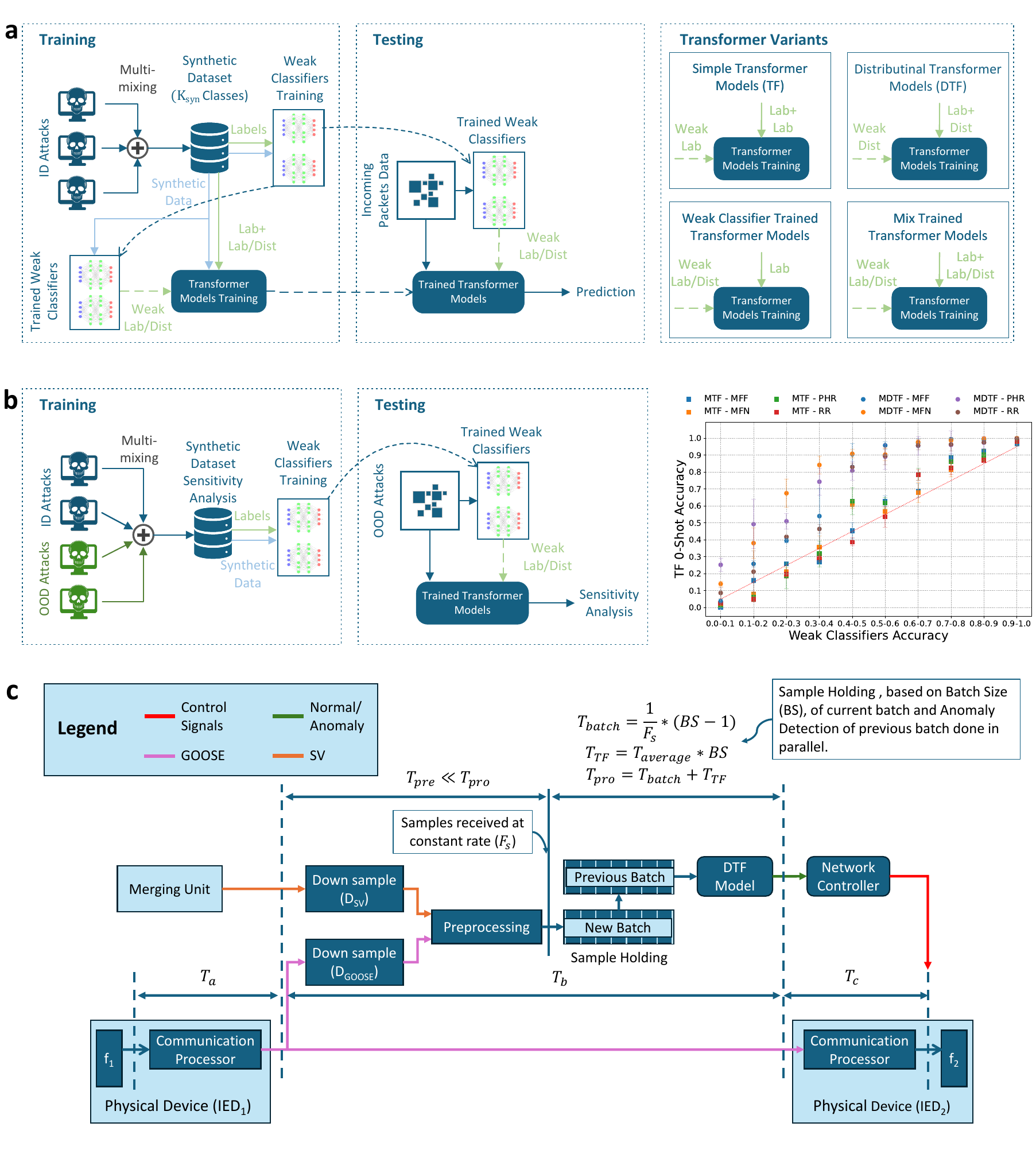}
    \vspace{-0.2cm}
    \caption{\textbf{High-level overview of experimentation frameworks.} \textbf{a}, Schematic of the overall training and testing methodology (refer to Figure \ref{fig:pretraining} for detailed architecture). Simple Transformer Models (TF) utilize hard labels (Lab) for in-context data, while Distributional Transformer Models (DTF) leverage distributions (Dist) for in-context data. Weak-classifier-trained transformers rely solely on weak Lab/Dist, whereas mixed-trained transformers incorporate both weak and ground truth Lab/Dist for in-context data. In Lab + X, Lab represent ground truth hard labels used for loss calculation and X represent ground truth Lab/Dist used in in-context data. 
\textbf{b}, Sensitivity analysis is conducted by synthetically training weak classifiers using in-distribution (ID) and out-of-distribution (OOD) classes. The pre-trained transformer model is then employed to evaluate OOD detection performance on these weak classifiers.  
\textbf{c}, Application of the proposed methodology in a real-world substation. $IED_1$ transmits GOOSE messages to $IED_2$ within the stringent latency requirement of 3 ms (per IEC-61850). Concurrently, sampled values (SV) from the merging unit and GOOSE packets are captured, downsampled ($D_{SV}$, $D_{GOOSE}$), preprocessed, and batched (batch size $BS$) before being processed by the proposed transformer-based detection framework (DTF Model). Each packet is processed with an average latency of $T_{average}(H,BS)$ seconds. Anomalous packets trigger a signal to the network controller, preventing the transmission of compromised messages.}
    \label{fig:experimentation}
\end{figure}
\section{Results}
\label{sec:experiments}
We validate the proposed ICL-based IDS framework on a real-world attack dataset from ERENO–IEC–61850 \cite{quincozes2023ereno}. The ERENO-IEC-61850 dataset provides a rich testing environment with nine distinct attack scenarios across two critical protocols (GOOSE and SV), effectively serving as multiple specialized datasets for different attack vectors. This diversity allows thorough evaluation across various attack patterns and network conditions. Specifically, the dataset contains seven types of GOOSE-based attacks, i.e., random and inverse replay, masquerade fake normal, masquerade fake attack, message injection, high-status number, and high-rate flooding. Additionally, it includes two types of SV-based attacks, i.e., message injection and inverse replay. Through the experiments, we aim to answer the following questions: 
\textbf{Q1:} How does increasing training attack diversity improve zero-day attack detection in digital substations? 
\textbf{Q2:}  What is the most effective way to leverage weak classifiers and ground-truth labels during training to enhance the model's performance?
\textbf{Q3:}  How is the model's performance affected by the weak classifier's performance?
\textbf{Q4:}  How well can the proposed approach detect zero-day attacks?
\textbf{Q5:}  Can transformer models be deployed in substations and adhere to the timing limitations of IEC-61850 protocol?

We utilize GPT-2 transformer architecture, with an in-context sample size of $N = 11$, for our experiments \cite{radford2019language}. For training, we choose 5 attacks: 3 from the GOOSE and 2 from the SV, along with the normal data. We treat these 5 selected attacks and the normal data as ID data, while all the other data from the ERENO dataset as OOD, which we use to test the zero-day attack detection accuracy.

Sections \ref{Q1} through \ref{Q5} address Questions 1 to 5, respectively. Refer to Figure \ref{fig:experimentation} for a visual representation of the experimentation frameworks.

\begin{figure}[p]
    \centering
    \includegraphics[width=1\textwidth]{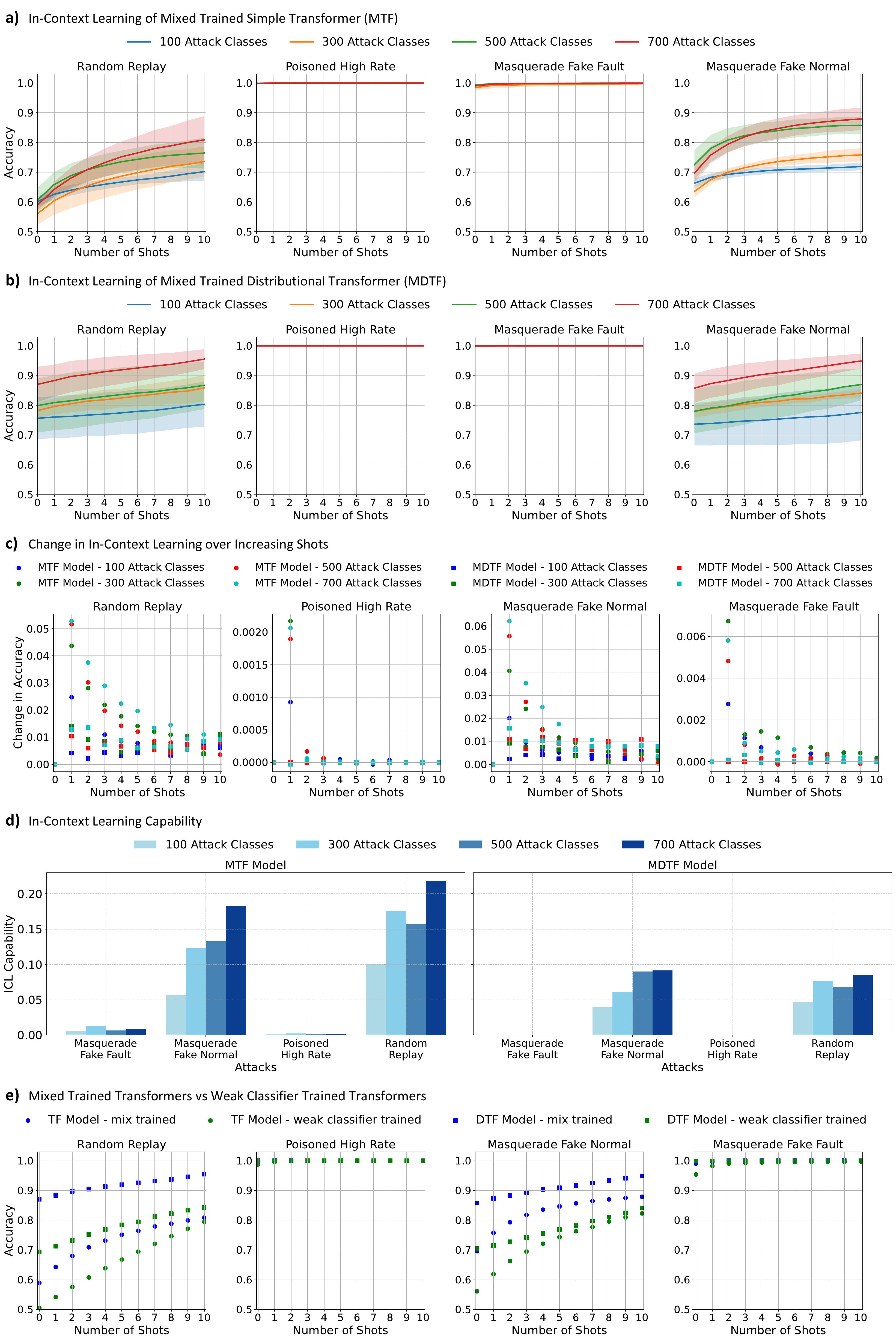}
    \caption{\textbf{Transformer model architectures perform under different training regimes and ablation studies.} \textbf{a}, In-context learning (ICL) performance of the mixed-trained distributional transformer (MDTF) under varying shot counts for out-of-distribution (OOD) attacks and different levels of attack-class diversity. Higher shot numbers consistently improve accuracy, underscoring the emergence of ICL across five random seeds.
\textbf{b} ICL performance of the mixed-trained transformer (MTF) model under the same conditions. As with MDTF, greater shot counts lead to higher accuracy, indicating robust ICL emergence over multiple seeds and training diversities.
\textbf{c}, Change in performance of MTF and MDTF across varying shot counts for OOD attacks, evaluated at different levels of training-attack diversity. MTF exhibits a more pronounced accuracy gain at lower shot numbers, indicating faster adaptation to OOD patterns compared to MDTF.
\textbf{d}, ICL capability of MTF and MDTF across OOD attacks for varying training diversity. Increasing the number of attack classes in training boosts ICL, with MTF generally surpassing MDTF under similar diversity conditions.
\textbf{e} Comparison of simple transformer (TF) and distributional transformer (DTF) models under mixed-trained and weak-classifier-trained strategies for various OOD attacks. Weak-classifier-trained models exhibit stronger ICL gains but reduced zero-shot performance relative to mixed-trained models.}
    \label{fig:Main_Results}
\end{figure}

\vspace{-0.2cm}
\subsection{The Impact of Training Data Diversity}\label{Q1}

To address Q1, we examine how training data diversity impacts ICL and zero-day attack detection performance by training the GPT-2 transformer on various numbers of attack classes (data diversity), \( K_{syn} \in \{100, 300, 500, 700\} \), generated using the multi-mixing approach described in our methodology in Section \ref{subsec:DataGeneration}. To quantify the performance of the transformer models on zero-day attack detection, we use the concept of \emph{shots}, which are the maximum instances of attacks in the in-context data. We quantify ICL capability as the total change in accuracy across maximum shots.

First, as shown in Figures \ref{fig:Main_Results}\textcolor{blue}{a} and \ref{fig:Main_Results}\textcolor{blue}{b}, increasing the number of shots (instances of attacks in the in-context data) improves accuracy across all four types of attacks not encountered during training for both the TF and DTF models. This trend indicates that the models are effectively adapting to OOD attack patterns as more attack packets are sent by attackers.

Second, we observe that the 0-shot accuracy (when the model encounters an OOD attack for the first time) of the DTF model is higher compared to the TF model. This suggests that DTF is better suited for scenarios where the volume of attack packets sent by an attacker is uncertain, as it can detect OOD attacks effectively even on the first instance. This characteristic makes DTF more effective in environments like digital substations, where even a single undetected attack packet could disrupt the whole system.

Third, as shown in Figure \ref{fig:Main_Results}\textcolor{blue}{c}, increasing training attack classes enhances the ICL capability for both TF and DTF models across different OOD attacks. For attacks like masquerade fake fault and poisoned high rate, the ICL capability is nearly zero, as the models achieve over 99\% accuracy on these attacks in a 0-shot scenario. Note that TF exhibits higher ICL capability across all tasks and classes than DTF, indicating superior generalization to OOD attacks. Furthermore, Figure \ref{fig:Main_Results}\textcolor{blue}{d} shows that TF demonstrates a higher rate of accuracy improvement per shot when shot numbers are low, meaning it can adapt to OOD attack patterns faster, requiring fewer samples for effective learning. In contrast, DTF has a lower and more constant adaptation rate. These results suggest that TF is generally more effective than DTF at generalizing and adapting to OOD attacks. 

These experiments were conducted for WCTF, MTF, WCDTF, and MDTF, all of which displayed similar trends of improved performance with an increased number of shots and greater training data diversity. This indicates that transformer models trained on a wider diversity of attacks are more effective at zero-shot OOD attack detection and adapt more quickly in n-shot settings for intrusion detection in digital substations. As the training dataset with 700 attack classes yielded the best performance, the following experiments are reported for this case only.

\vspace{-0.1cm}
\subsection{Comparative Analysis of Label Choices During Training}\label{subsec:GTvsWC}

To address Q2, we examine the impact of different in-context Lab/Dist choices during training: (1) Lab/Dist from weak classifiers only and (2) a mixture of weak classifier and ground-truth Lab/Dist.

First, we experimented with various mixing ratios. 
We discovered that for the TF models, a training ratio of 60\% weak classifier labels and 40\% ground-truth labels yielded the highest attack detection accuracy during testing. In contrast, for the DTF model, we achieved the best performance with a minimal ground-truth label ratio of 5\%. This asymmetry appears to stem from the different information density in label versus distribution-based training. Additionally, we found that using a high proportion of ground-truth labels  (greater than 10\%), in the case of DTF, led to model instability, characterized by decreasing accuracy with additional shots and significant performance variation across five different seeds.

Figure \ref{fig:Main_Results}\textcolor{blue}{e} compares attack detection accuracy during testing for TF and DTF models trained with only weak classifier labels versus mixed labels (40:60 for TF and 95:5 for DTF). The figure clearly illustrates that incorporating a mix of ground truth and weak classifier labels during training improves both zero-shot and n-shot performance.

The WCTF model shows low zero-shot performance but high ICL, while the MTF model demonstrates both high zero-shot performance and high ICL, suggesting that including ground-truth labels for in-context data during training enhances the transformer models' attack detection accuracy. From an engineering perspective, DTF offers an advantage over TF, as a stable mix of only 2–5\% ground-truth labels yields the best results, eliminating the need to fine-tune the ground-truth-to-weak-classifier label ratio and reducing the number of hyper-parameters requiring adjustment.

\subsection{Sensitivity Analysis}\label{Q4}
To address Q3, we conducted a sensitivity analysis of our models by synthetically training multiple weak classifiers with accuracies varying in increments of 0.1 for OOD attacks. We adjusted the ratio of normal-to-attack instances and included some OOD attacks as in-distribution during the training of these weak classifiers. The zero-shot performance of MDTF and MTF models was then evaluated across a task diversity of 700 using all sets of weak classifiers.

Figure \ref{fig:experimentation}\textcolor{blue}{b} illustrates the relationship between the accuracy of weak classifiers (calculated using hard-label ensembling technique) and the zero-shot performance of the transformer models. The analysis shows that as the accuracy of weak classifiers increases, the performance of both MDTF and MTF models improves. This trend is expected because as more weak classifiers start predicting correctly, the transformer models start to give more weightage to their cumulative response.

Note that MDTF consistently outperforms weak classifiers, while MTF surpasses weak classifiers once they achieve an accuracy threshold of 0.5–0.6. This dependence of MTF on classifier accuracy is a potential limitation, as ideally, the models should outperform weak classifiers across all accuracy ranges.

We also observe a distinct pattern in the performance curves: MTF exhibits a linear relationship between weak classifier accuracy and zero-shot performance, while MDTF displays a parabolic relationship. This means that even with low ensembling accuracy (0.4–0.5), MDTF is able to achieve high accuracy (over 0.8) in detecting OOD attacks. However, the 0.4-0.5 range represents a critical threshold, as lower classifier accuracies cause a sharp decline in MDTF’s performance.

In real-world applications, weak classifiers often perform unpredictably on OOD attacks, but MDTF proves highly reliable, achieving strong detection accuracy even when these classifiers operate as low as 0.4–0.5. This robustness ensures MDTF remains effective in uncertain environments, making it a valuable approach for zero-day attack detection.

\definecolor{highlight}{rgb}{0.96, 0.76, 0.76} 


\begin{sidewaystable*}[h]
\caption{Performance of different models on out-of-distribution (OOD) and in-distribution (ID) data.}
\label{tbl:MainTable}
\resizebox{\textwidth}{!}{%
\begin{tabular}{@{\extracolsep{\fill}}lcccccccccc}
\toprule
& \multicolumn{4}{@{}c@{}}{\textbf{OOD}} 
& \multicolumn{6}{@{}c@{}}{\textbf{ID}} \\
\cmidrule(lr){2-5}\cmidrule(lr){6-11}
\textbf{Models} 
& \makecell{\textbf{Poisoned} \\ \textbf{High Rate}} 
& \makecell{\textbf{Masquerade} \\ \textbf{Fake Fault}} 
& \makecell{\textbf{Masquerade} \\ \textbf{Fake Normal}} 
& \makecell{\textbf{Random} \\ \textbf{Replay}} 
& \textbf{Normal} 
& \makecell{\textbf{High} \\ \textbf{Status Number}} 
& \makecell{\textbf{Inverse} \\ \textbf{Replay}} 
& \textbf{Injection}
& \makecell{\textbf{SV High} \\ \textbf{Status Number}}
& \makecell{\textbf{SV} \\ \textbf{Injection}} \\
\midrule

Logistic Regression 
& 1.000 & 1.000 & \cellcolor{highlight}0.102 & 1.000 
& 1.000 & 1.000 & 1.000 & 1.000 
& 1.000 & 1.000 \\

Decision Tree 
& 1.000 & 1.000 & \cellcolor{highlight}0.000 & 1.000 
& 1.000 & 1.000 & 1.000 & 1.000 
& 1.000 & 1.000 \\

Random Forest 
& 1.000 & 0.859 & \cellcolor{highlight}0.008 & 0.993 
& 1.000 & 1.000 & 1.000 & 1.000 
& 1.000 & 1.000 \\

SVM 
& 1.000 & 0.996 & \cellcolor{highlight}0.091 & 1.000 
& \cellcolor{highlight}0.970 & 1.000 & 1.000 & 1.000 
& 1.000 & 1.000 \\

Naive Bayes 
& 0.976 & \cellcolor{highlight}0.410 & \cellcolor{highlight}0.721 & \cellcolor{highlight}0.000 
& \cellcolor{highlight}0.934 & 1.000 & 0.814 & \cellcolor{highlight}0.280 
& 1.000 & 1.000 \\

DNN 
& \cellcolor{highlight}0 & \cellcolor{highlight}0.025 & 0.968 & \cellcolor{highlight}0.007
& \cellcolor{highlight}0.999 & 1.000 & \cellcolor{highlight}0.333 & \cellcolor{highlight}0.636
& 1.000 & \cellcolor{highlight}0.788 \\

CNN 
& 0.997 & \cellcolor{highlight}0.400 & 0.804 & \cellcolor{highlight}0.288
& \cellcolor{highlight}0.932 & 1.000 & 0.935 & \cellcolor{highlight}0.765 
& 1.000 & 1.000 \\

RNN 
& 1.000 & 1.000 & \cellcolor{highlight}0 & 1.000 
& 1.000 & 1.000 & 1.000 & 1.000
& 1.000 & 1.000 \\

LSTM 
& 1.000 & 1.000 & \cellcolor{highlight}0 & 1.000
& 1.000 & 1.000 & 1.000 & 1.000 
& 1.000 & 1.000 \\

Hard Voting WC 
& 1.000 & 0.999 & \cellcolor{highlight}0.608 & \cellcolor{highlight}0.517 
& 1.000 & 0.999 & 0.997 & 1.000 
& 0.978 & 1.000 \\

MTF 0-Shot \textbf{(Ours)} 
& 0.998 & 0.991 & \cellcolor{highlight}0.696 & \cellcolor{highlight}0.590 
& 1.000 & 1.000 & 1.000 & 1.000 
& 1.000 & 1.000 \\

MTF Max-Shot \textbf{(Ours)} 
& 1.000 & 0.999 & 0.879 & 0.809 
& 1.000 & 1.000 & 1.000 & 1.000 
& 1.000 & 1.000 \\

MDTF 0-Shot \textbf{(Ours)} 
& 1.000 & 1.000 & 0.858 & 0.871 
& 1.000 & 1.000 & 1.000 & 1.000 
& 1.000 & 1.000 \\

MDTF Max-Shot \textbf{(Ours)} 
& 1.000 & 1.000 & 0.949 & 0.956 
& 1.000 & 1.000 & 1.000 & 1.000 
& 1.000 & 1.000 \\

\botrule
\end{tabular}
}
\end{sidewaystable*}

\subsection{Zero-day Attack Detection}\label{AA}
We answer Q4 by validating the trained model on the attacks not seen during the training (OOD attacks) and on the attacks that are seen during the training (ID attacks). We compare our results against widely known ML-based classification methods. The TF and DTF models are trained across 700 attack classes, using a 40:60 of ground-truth to weak classifiers Lab/Dists for TF, and a 5:95 for DTF.

Table \ref{tbl:MainTable} presents the final results, where red cells indicate 'failure cases,' defined as: \emph{1)} inability to reach an accuracy threshold of 80\% for a specific attack, and \emph{2)} failure to achieve 100\% accuracy on normal data.

As shown in Table \ref{tbl:MainTable}, widely used ML-based methods encountered at least one failure case, demonstrating limitations in detecting certain zero-day attacks. Furthermore, traditional ensembling techniques, like hard voting applied to weak classifiers, also resulted in failure cases. In contrast, our proposed ICL-based methods, evaluated under max-shot conditions, did not show these failures when using weak classifiers. Although the MTF model encountered a failure case in the 0-shot scenario, it still achieved higher accuracy than hard-voting accuracy with weak classifiers. The MDTF model, however, demonstrated no failure cases, even in the 0-shot scenario, and surpassed 90\% accuracy under max-shot conditions across all OOD attacks. This observation suggests an inherent capacity of MDTF models to recognize previously unseen attacks.

For known attacks, both MDTF and MTF models achieved greater than 95\% accuracy under both 0-shot and max-shot scenarios, indicating their comparable performance to traditional ML models in detecting known threats. Additionally, achieving 100\% accuracy on normal data signifies that our proposed models had zero false positives. This is crucial in digital substations, as minimizing false alarms is essential to avoid unnecessary disruptions to operations.


\begin{sidewaystable*}[h]
\caption{Average inference time per sample ($T_{average}$) under various batch sizes ($BS)$, and hardware and frameworks ($H)$.}
\label{tab:timing}
{
\begin{tabular*}{\textwidth}{@{\extracolsep{\fill}}lccccc}
\toprule
{\makecell{Batch \\ Size}} 
& {\makecell{CPU \\ (Original) \\ (s)}} 
& {\makecell{CPU \\ (ONNX) \\ (s)}} 
& {\makecell{CPU \\ (OpenVINO) \\ (s)}} 
& {\makecell{GPU \\ (Original) \\ (s)}}  
& {\makecell{GPU \\ (TensorRT) \\ (s)}} \\ 
\midrule
1   & $0.09931\pm0.01742$ & $0.02986\pm0.00421$ & $0.03560\pm0.01477$ & $0.01700\pm0.00418$ & $0.00227\pm0.00030$ \\ 
10  & $0.03413\pm0.00171$ & $0.01154\pm0.00023$ & $0.01044\pm0.00134$ & $0.00173\pm0.00048$ & $0.00035\pm0.00014$ \\ 
20  & $0.03494\pm0.00557$ & $0.01148\pm0.00098$ & $0.01084\pm0.00205$ & $0.00106\pm0.00021$ & $0.00018\pm0.00015$ \\ 
100 & $0.02942\pm0.00315$ & $0.01025\pm0.00063$ & $0.00742\pm0.00069$ & $0.00077\pm0.00021$ & $0.00005\pm0.00003$ \\ 

\bottomrule
\end{tabular*}
}
\end{sidewaystable*}

\subsection{Deployment in Digital Substations}\label{Q5}
To answer Q5 and demonstrate the feasibility of deploying our DTF model for real‐time anomaly detection in IEC 61850‐compliant digital substations, we trained the model offline on an NVIDIA RTX A6000 GPU (5000 iterations, ~1 hour of training). After training, we evaluated inference performance on standard computing platforms—an AMD Ryzen 7 6800H CPU and an NVIDIA RTX 3070 Ti GPU—using different optimization and runtime frameworks (ONNX, OpenVINO, TensorRT). Table \ref{tab:timing} reports the average inference time per sample under various batch sizes, hardware, and frameworks.


Figure~\ref{fig:experimentation}\textcolor{blue}{c} illustrates the high‐level integration of our DTF model in a digital substation. At the process‐bus level, SV packets arrive from MUs at a sampling rate \(F_{SV} = 4800\,{Hz}\), while GOOSE packets arrive from IEDs at a rate \(F_{GOOSE}\). Under normal operating conditions, \(F_{GOOSE}\) may be as low as \(0.5\,{Hz}\), but it can surge to \(250\,{Hz}\) during faults \cite{Hou2010IEC6}. Both SV and GOOSE streams are downsampled by factors \(D_{SV}\) and \(D_{GOOSE}\), respectively, and then preprocessed. Thus, the combined data rate at the input of the model is 
\[
F_{s} \;=\; \max \bigl( F_{GOOSE}\times D_{GOOSE}, \; F_{SV}\times D_{SV} \bigr).
\]
To improve inference throughput, we collect data in batches of size \(BS\). Once \(BS\) preprocessed samples are accumulated, the batch is fed to the model. Let \(T_{average}(H,BS)\) be the average per‐sample inference time on hardware and framework \(H\) and batch size \(BS\). Then, the time to process one full batch is
\[
T_{TF} \;=\; T_{average}(H,BS)\,\times\,(BS-1), 
\quad
T_{batch} \;=\; \frac{BS}{F_{s}},
\]
\[
T_{pro} \;=\; T_{batch} + \;T_{TF},
\]
where \(T_{pro}\) represents the total time overhead introduced by batching and processing. The total detection latency is then 
\[
T_{total} \;=\; T_{pre} + T_{pro},
\]
where \(T_{pre}\) is the (typically small) preprocessing time.


IEC 61850 classifies substation applications based on how quickly messages (e.g., GOOSE) must be exchanged among networked IEDs. For instance, the fastest category (Type 1A, performance class P2/P3) specifies a 3 ms upper bound for message “transmission time” (\(T_{transmission}\)) in fast‐messaging (trip) applications \cite{Hou2010IEC6}. The standard further decomposes \(T_{transmission}\) into:
1. \(T_{a}\): the time within the sending IED’s communications processor.
2. \(T_{b}\): the physical transit time of the message across the substation network.
3. \(T_{c}\): the time within the receiving IED’s communications processor.

Although IEC 61850 does not place strict constraints on \(T_{a}\), \(T_{b}\), or \(T_{c}\) individually, it mandates that the total \(T_{transmission}\) not exceed the standard’s specified limit (3 ms for Type 1A). To align our model’s inference with this performance class, we make the following simplifying assumptions:
\begin{enumerate}
    \item \(T_{pre} \ll T_{pro}\): Preprocessing time is negligible compared to batching and model inference time, so we approximate \(T_{total} \approx T_{pro}\).
    \item \(T_{b} \gg T_{a} + T_{c}\): The transit time across the network dominates the communication‐processor latencies, thus \(T_{transmission} \approx T_{b}\).
    \item The Network Controller can interrupt an operation immediately upon detecting an anomalous packet.
\end{enumerate}

Under these assumptions, our constraint for real‐time anomaly detection becomes:
\[
T_{total} \;<\; T_{transmission},  \quad T_{average}(H,BS) \;<\; \frac{1}{F_{s}}.
\]
Effectively, the DTF model must produce an anomaly (or normal) decision before a high‐priority message (e.g., GOOSE trip) completes its propagation through the substation network and before the next preprocessed sample arrives.


\[
\mathop{\arg\min}\limits_{BS,\,H,\,D_{{GOOSE}},\,D_{{SV}}}
T_{average}(H,BS)\footnotemark[1]
\]
\[
\text{subject to} \quad
T_{total}\footnotemark[2] < T_{transmission}, 
\quad
T_{average}(H,BS) < \frac{1}{F_{s}\footnotemark[3]}.
\]

\footnotetext[1]{\(T_{average}(H,BS)\) depends on the chosen hardware \(H\) and batch size \(BS\).}
\footnotetext[2]{\(T_{total}\) depends on \(T_{pro}\), which in turn depends on \(T_{TF}\) and \(T_{batch}\); these depend on $T_{average}$,\(BS\) and \(F_s\).}
\footnotetext[3]{\(F_s\) depends on \(F_{GOOSE}\), \(D_{{GOOSE}}\), \(F_{SV}\), and \(D_{{SV}}\).}

For fault conditions, we consider \(F_{GOOSE} = 250\,{Hz}\), \(F_{SV} = 4800\,{Hz}\), and \(T_{transmission} = 3\,{ms}\). From Table \ref{tab:timing}, the $H$ = GPU‐based TensorRT framework yields a per‐sample inference time \(T_{average}(H,BS) = 2.44\,{ms}\) that satisfies \(T_{total} < 3\,{ms}\) for \(BS = 1\). Our analysis shows that \(D_{GOOSE} = 1\) and \(D_{SV} = \tfrac{1}{12}\) suffice to ensure \(T_{average}(H,BS) < \tfrac{1}{F_s}\), where \({F_s} = {F_{SV}} \times {D_{SV}}\). Thus, the model can detect anomalies in time to prevent spurious or unintended operations even in the time‐sensitive IEC 61850 applications, all while using a cost‐effective GPU.

\section{Discussion}
Here we propose two distinct transformer architectures that harness ICL to detect zero-
day attacks in digital substations. In doing so, we also demonstrate how inaccurate or
noisy labels in in-context can be leveraged to enhance generalization. Through sensitivity
analysis we show that weak learners achieving only 40–50\% accuracy can drive the transformers
to attain at least 80\% detection accuracy. Our approach is validated via extensive
experiments on the publicly available IEC-61850 attack dataset, comparing favorably
against widely used ML-based baselines for zero-day attack detection. Finally, we
illustrate how these architectures can be deployed in a real substation environment that
adheres to the time senstive IEC-61850 protocol, underscoring the practical viability
of our solution.

\textbf{Limitations and Future Directions.}
Although our models have capability of performing multi-classification, they currently lack explicit mechanisms to distinguish in-distribution (ID) from out-of-distribution (OOD) attacks. Enabling a more nuanced differentiation could help substation engineers implement targeted countermeasures. Additionally, while our sensitivity analysis focuses on weak classifier accuracy as a critical external factor, performance may vary if weak classifiers dip below the 40–50\% accuracy range. Finally, exploring other parameters such as sequence length and architectural variants could further refine detection accuracy and robustness.

\begin{figure*}[!ht]
    \centering
\includegraphics[width=1\textwidth]{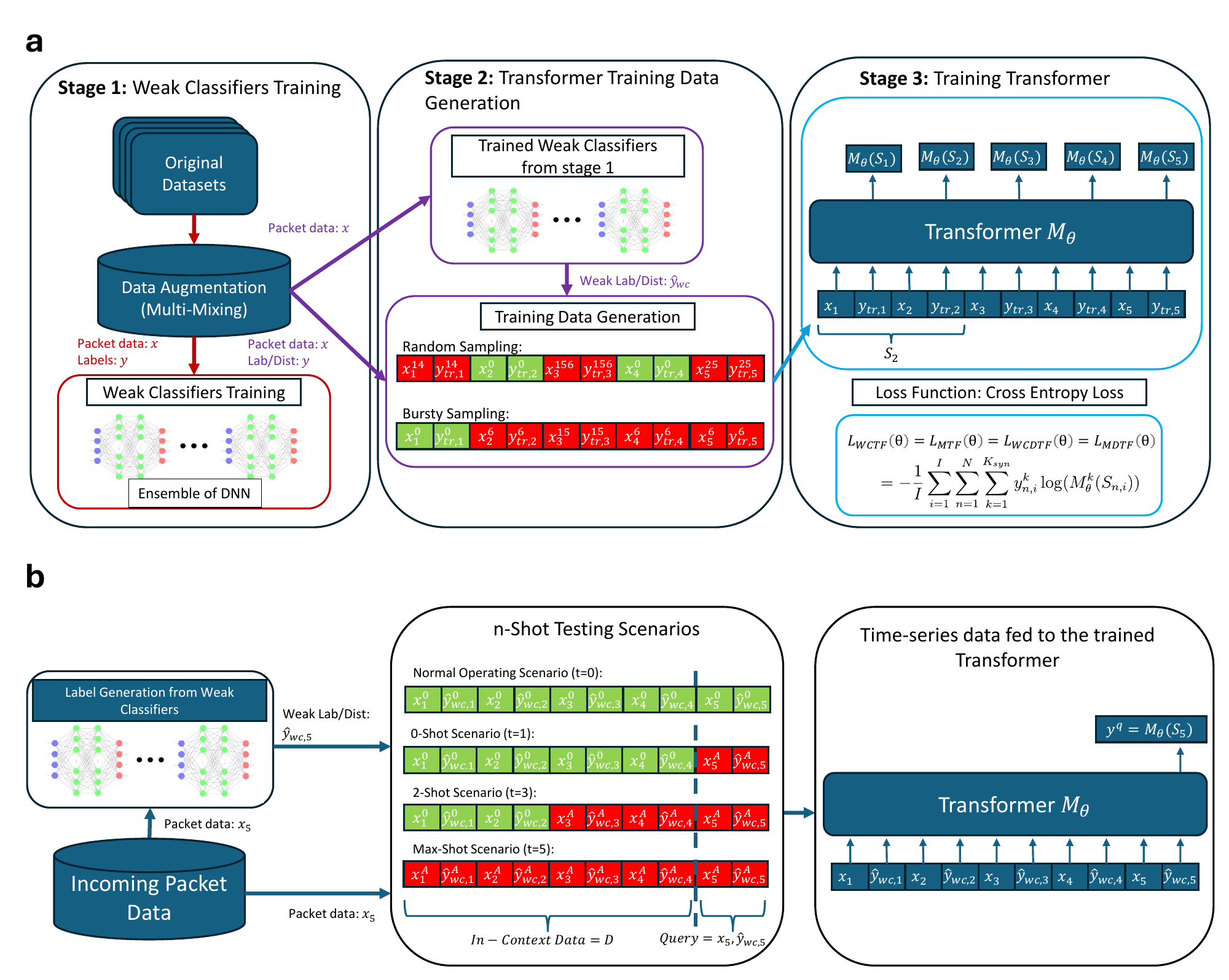}
    \vspace{-0.2cm}
    \caption{\textbf{Transformer training (a) and testing (b) frameworks}. \textbf{a}, In stage 1, synthetic data is generated using multi-mixing. Using this data, weak classifiers are trained. In stage 2, data samples $(x_1,{y}_{tr,1},\dots, x_5,{y}_{tr,5})$ are generated for training, where $x_i$ represents packets from synthetic data and ${y}_{tr,i}$ represents their labels/distributions (Lab/Dist) that can be from weak classifier or mixture of weak classifiers and ground truth. Finally, in stage 3, the transformer is trained with cross-entropy loss, where the loss function is same for Weak Classifier Trained Transformer (WCTF), Mixture of weak classifiers and ground-truth Trained Transformer (MTF), Weak Classifier Trained Distributional Transformer (WCDTF), and Mixture of weak classifiers and ground-truth Trained Distributional Transformer (MDTF). \textbf{b}, Incoming packet $(x_5)$ is received and passed to pre-trained weak classifiers to generate weak labels/distributions $(\hat{y}_{wc,5})$. This data-label pair query $(x^{qs} = (x_5,\hat{y}_{wc,5}))$ is then appended at the end of the sequence. The trained transformer model sees different scenarios depending on the incoming packets in the in-context data and then predicts the label $y^q$ for $x_5$ using ICL.}

    \label{fig:pretraining}
\end{figure*}

\section{Methodology}
In this section, we describe our overall ICL-based IDS framework with the training and testing procedures in detail. 

\label{sec:Methodology}
\subsection{Training Data Generation}\label{subsec:DataGeneration}
Many recent works have highlighted the importance of training data diversity to improve ICL capabilities in transformer models \cite{raventos2024pretraining,min2021metaicl,levy2023diversedemonstrationsimproveincontext}. Specifically, within the cybersecurity context, this translates to the inclusion of various types of attack scenarios. However, there are not many existing cyber-attack datasets for the IEC-61850 protocol.
Therefore, to increase the diversity of the training data for ICL capabilities, we propose an approach called \emph{multi-mixing}, where the key idea is to generate synthetic attacks, without collecting additional data, by linearly combining features from different attack classes. By creating these synthetic examples, we aim to expose the transformer model to a wider range of attack patterns, potentially improving its ability to detect zero-day attacks.
In the traditional mixup approach \cite{zhang2018mixup}, two data points are selected randomly and combined according to weights derived from a beta distribution. Instead, in our proposed multi-mixing strategy, we extend this concept by selecting random data points from each available class and linearly combining them to form a new, synthetic class. This approach ensures each new data point is a mixture of multiple classes, enhancing the dataset's overall diversity.
To formalize this, consider a dataset with $K_{ori}$ distinct classes. Multi-mixing generates a new synthetic class as follows: 
\begin{equation}
\label{eq:multimixing}
A_{\text{new}} = \sum_{k=1}^{K_{ori}} \alpha_k A_k,
\end{equation}
where $A_{k}$ represents all data points (shuffled) belonging to the $k^{th}$ class, and $\alpha_k \in [0, 1]$ denotes the weight of the $k^{th}$ class incorporated into the new class, denoted as $A_{new}$. Each of these new classes is assigned a new label, which allows the model to learn more latent features and generalize better. We denote the total number of classes generated after applying the multi-mixing process as $K_{syn}$, which includes the original classes ($K_{ori}$) and represents the training attack diversity.

 Algorithm \ref{alg:generate_attacks_with_repetitions} describes the procedure to perform multi-mixing. This approach automatically assigns values to mixing coefficients $\alpha_k$ during the normalization (see line 6).
\begin{algorithm}
\caption{Multi-mixing Algorithm}
\label{alg:generate_attacks_with_repetitions}
\begin{algorithmic}[1]
\Require $total\_attack\_classes$, $known\_attacks$
\Ensure $new\_attacks$, $labels$
\State Initialize $new\_attacks \Leftarrow \{\}$ \Comment{Empty list for new attacks}
\State Initialize $labels \Leftarrow \{\}$ \Comment{Empty list for labels}
\State Set $k \Leftarrow 1$, $label \Leftarrow 1$, $attack\_NO \Leftarrow 0$

\While{$attack\_NO < total\_attack\_classes$}
    \ForAll{$attack\_mixture \in combinations\_with\_replacement(known\_attacks, k)$}
        \State $new\_attack \Leftarrow \frac{\sum attack\_mixture}{len(attack\_mixture)}$
        \State Append $new\_attack$ to $new\_attacks$
        \State Append $label$ to $labels$
        \State $label \Leftarrow label + 1$
        \State $attack\_NO \Leftarrow attack\_NO + 1$
        \If{$attack\_NO == total\_attack\_classes$}
            \State \textbf{break}
        \EndIf
    \EndFor
    \State $k \Leftarrow k + 1$
\EndWhile

\Return $new\_attacks, labels$
\end{algorithmic}
\end{algorithm}

\subsection{In-context Data from Weak Classifiers}\label{subsec:WeakClassifier}
In our proposed methodology, the transformer model uses ICL to detect the novel incoming attack packet as normal or attack. However, during testing, the true labels of incoming data packets are not available. To achieve ICL, the in-context data should consist of the input-label pairs (same as $D_{N-1}$ in Section \ref{subsec:ICL}). Hence, we utilize weak classifiers, which are pre-trained models (such as neural networks). These models generate probability scores, \( y_{pr,n}^k \in [0, 1] \), representing the likelihood that the \( n^{th} \) data packet, \( x_n \in \mathbb{R}^d \), in the in-context data belongs to class \( k \in \{0, 1, \ldots, K_{syn}\} \). The class with the highest probability, \( \hat{y}_n = \arg\max_k y_{pr,n}^k \), is then used as a pseudo-label for the in-context data processed by the transformer model.

These classifiers are termed "weak" because they may not achieve perfect detection accuracy but offer preliminary predictions that guide the transformer model’s ICL process. To reduce the influence of individual weak classifier errors, we aggregate the argmax outputs from multiple weak classifiers and provide the concatenated results as input to our transformer model as follows:
\[
y_{wc,n} = [\hat{y}_n^1 ,\hat{y}_n^2, \ldots, \hat{y}_n^W, \boldsymbol{-1}] \in \mathbb{R}^d,
\]
where \( \hat{y}_n^w \) denotes the class predicted by the \( w^{th} \) weak classifier for the \( n^{th} \) data packet in in-context data. The padding token \( \boldsymbol{-1} \) is used to ensure uniform input length, and \( W \) is the number of weak classifiers. We employ deep neural networks as weak classifiers, designed specifically for multi-class classification.

\subsection{Training the Transformer Model}\label{subsec:transformerTraining}
To train the transformer model for intrusion detection, we explore two possible labeling approaches for each in-context sample: \emph{1)} labels from weak classifiers, and \emph{2)} a combination of weak classifier labels and ground-truth labels. We label the former as Weak Classifier Trained Transformer (WCTF) and the latter as Mixed Trained Transformer (MTF). Our experimental results, presented in Section \ref{subsec:GTvsWC}, demonstrate that this mixed labeling approach achieves the highest accuracy for zero-day attack detection.

We train the transformer model using \( N \) input-label pairs, denoted as \( \{(x_1, y_{tr,1}), (x_2, y_{tr,2}), \ldots, (x_{N}, y_{tr,N})\} \), where \(\text{tr}\) indicates a training sample. Each \( x_n \) is sampled randomly and burstily from the \( K_{syn}-1 \) attack classes and the normal data \cite{chan2022data}. The label \( y_{tr,n} \) for each sample \( n \in \{1, 2, \ldots, N\} \) is determined by the following labeling methods: when using only weak classifier labels, the \( y_{tr,n} \) is denoted by \( {y}_{wc,n} \); when using mixed-labels, it is represented as \( y_{mix,n} \).

For each input \( n \), the transformer model considers the data \( S_n = (x_1, y_{tr,1}, \ldots, x_n, y_{tr,n}) \)—where \( D_{n-1} = \{(x_1, y_{tr,1}), (x_2, y_{tr,2}), \ldots, (x_{n-1}, y_{tr,n-1})\} \) denotes the in-context data and $(x_n, y_{tr,n})$ denotes the query set, \( x^{qs} = (x_n, y_{tr,n}) \)— to make its prediction \( M_{\theta}(S_n) \) for target \( y_n \). We append \( y_{tr,n} \) alongside \( x_n \) to provide additional information to the model about the potential true label.

We use the cross-entropy loss functions for training: \emph{(i)} WCTF and \emph{(ii)} MTF loss:

\begin{equation}
    L_{WCTF}(\theta) = L_{MTF}(\theta) =
-\frac{1}{I} \sum_{i=1}^{I} \sum_{n=1}^{N} \sum_{k=1}^{K_{syn}} 
y_{n,i}^k \log(M_{\theta}^{k}(S_{n,i})),
\end{equation}

where \( M_{\theta}^{k}(\cdot) \) denotes the probability of \( x_n \) in ${i^{th}}$ data sample belonging to class \( k \), and \( y_{n,i}^k \) is an indicator of the correct label for \( x_n \) in ${i^{th}}$ data sample. Although we initially train the model for multi-class classification, our primary goal is binary classification, specifically detecting whether incoming packets are normal or represent an attack.

\subsection{Distributional Transformer Model}\label{subsec:Distributional_Transformer}

In Section \ref{subsec:WeakClassifier}, we generated weak classifiers' outputs \( y_{wc,n} \) for in-context data by concatenating \( \hat{y}_n^w \), the hard-label prediction from each \( w^{th} \) weak classifier for the \( n^{th} \) data packet in in-context data. Here, we consider a more nuanced approach: instead of using only hard-label predictions, we investigate whether incorporating the full probability distribution over $K_{syn}$ classes from each weak classifier could enhance zero-day attack detection.

We propose a distributional transformer (DTF) model, which takes modified weak classifiers' outputs, \( y_{wcd,n} \), defined as:

\[
{y}_{wcd,n} = [\hat{y}_{prd,n}^{1}, \hat{y}_{prd,n}^{2}, \ldots, \hat{y}_{prd,n}^{W}] \in \mathbb{R}^{K_{syn} \times W},
\]
\[
\hat{y}_{prd,n}^{w} = [{y}_{pr,n}^{w,1},{y}_{pr,n}^{w,2},\ldots,{y}_{pr,n}^{w,K_{syn}}] \in \mathbb{R}^{K_{syn}},
\]

where \( \hat{y}_{pr,n}^{w,k} \in [0, 1] \) represents the probability determined by \( w^{th} \) weak classifier for \( n^{th} \) data packet in $S_N$ to belong to \(k^{th}\) class. We refer to the transformer model introduced in Section \ref{subsec:transformerTraining} as the simple transformer (TF) for comparison. 

Similar to TF, DTF can be trained with two types of in-context probability distributions: (1) the weak classifier distribution and (2) a combination of the weak classifier and ground-truth distributions.

The training approach for DTF is similar to that of TF, but the model is now trained with \( N \) input-label pairs, \( \{(x_1, y_{tr,1}), (x_2, y_{tr,2}), \ldots, (x_N, y_{tr,N})\} \) 
, where \( y_{tr,n} = {y}_{wcd,n} \) if only the weak classifier probability distribution is used, and \( y_{tr,n} = y_{mixd,n} \) when using the mixed distribution. The following cross-entropy loss function is used to train DTF:

\begin{equation}
    L_{WCDTF}(\theta) = L_{MDTF}(\theta) = \\
-\frac{1}{I} \sum_{i=1}^{I} \sum_{n=1}^{N} \sum_{k=1}^{K_{syn}} 
y_{n,i}^k \log(M_{\theta}^{k}(S_{n,i})),
\end{equation}

with the final output used for binary classification.

\subsection{Testing}
\label{subsec:testing}
Once trained, we can deploy the transformer models to detect anomalies in real-time.
The most recent data packet received, and its weak classifier Lab/Dist are treated as the query set, $x^{qs} = (x_N,\hat{y}_{wc,N})$, where $\hat{y}_{wc,N} = {y}_{wc,N}$ for TF models and ${y}_{wcd,N}$ for DTF models. The preceding $N-1$ packets and their weak labels/distributions serve as the in-context data $D_{N-1}$. Under standard substation operations, we would anticipate that all packets within the sequence are normal. However, in the event of an attack, the $x_{N}$ becomes anomalous, while the earlier in-context packets would likely still reflect normal conditions. We refer to the transformer model's ability to detect anomalies where the context remains normal while only the query point is anomalous—as its \textbf{zero-shot} performance. This scenario assesses the model's ability to detect completely novel attacks based solely on its learned representations.

As the attack persists, the attacker continues to send anomalous packets, which begin to appear in the in-context data. The model's performance in this setting evaluates its ability to adapt and detect the new attack based on these few new unlabeled examples. This gradual transition from normal to anomalous in the in-context data allows us to evaluate the model's \textbf{n-shot} performance, where n denotes the number of anomalous packets present within the in-context data $D_{N-1}$.

Note that our approach differs from traditional ICL. Instead of having access to true labels, we rely solely on pseudo-labels generated by weak classifiers. From a deployment perspective, this methodology is especially beneficial for real-time intrusion detection in digital substations. It enables the model to quickly adapt to new types of attacks based on just a few observed instances. 


\vspace{-0.2cm}

\bibliography{references}

\begin{appendices}
\printnomenclature
\nomenclature[A]{$K_{ori}$}{Number of original classes used to generate synthetic classes via multi-mixing.}
\nomenclature[A]{$K_{syn}$}{Number of classes (including $K_{ori}$) after applying the multi-mixing process used in training transformer models.}
\nomenclature[A]{$N$}{Length of in-context data + 1 (length of query set).}
\nomenclature[A]{$I$}{Number of training samples used to train the transformer model.}
\nomenclature[A]{$W$}{Number of weak classifiers used to generate pseudo labels/distributions for the packet data.}

\nomenclature[B]{$D_{n}$}{$n$ input-label/distribution pairs representing in-context data.}
\nomenclature[B]{$S_{n}$}{$n$ input-label/distribution pairs representing both in-context data ($D_{n-1}$) and query set ($x^{qs}$).}
\nomenclature[B]{$x^{qs}$}{Query set (input-label/distribution pair) for which the transformer model predicts if the data packet is malicious or normal, given $D_{n}$.}

\nomenclature[C]{$\hat{y}_n^w$}{Hard label output from the $w^{th}$ weak classifier for the $n^{th}$ data packet sample in $S_{N}$.}
\nomenclature[C]{$y_{pr,n}^{w,k}$}{Probability that the $n^{th}$ data packet sample in $S_{N}$ belongs to class $k$ as predicted by the $w^{th}$ weak classifier.}
\nomenclature[C]{$\hat{y}_{prd,n}^{w}$}{Probability distribution across all $K_{syn}$ classes from the $w^{th}$ weak classifier for the $n^{th}$ data packet sample in $S_{N}$.}

\nomenclature[D]{$y_{wc,n}$}{Concatenated hard label outputs from all weak classifiers for the $n^{th}$ data packet sample in $S_{N}$.}
\nomenclature[D]{$y_{wcd,n}$}{Concatenated probability distributions from all weak classifiers for the $n^{th}$ data packet sample in $S_{N}$.}
\nomenclature[D]{$\hat{y}_{wc,n}$}{Concatenated hard label (for simple transformers) or distribution (for distributional transformers) outputs from all weak classifiers for the $n^{th}$ data packet sample in $S_{N}$.}

\nomenclature[E]{$y_{tr,n}$}{Concatenated labels from weak classifiers or a mixture of weak classifiers and ground truth labels/distributions for the $n^{th}$ data packet sample in $S_{N}$.}
\nomenclature[E]{$y_{mixd,n}$}{Concatenated combination of probability distributions from all weak classifiers and the true distribution for the $n^{th}$ data packet sample in $S_{N}$.}
\nomenclature[E]{$y_{mix,n}$}{Concatenated combination of labels from all weak classifiers and ground truth labels for the $n^{th}$ data packet sample in $S_{N}$.}
\nomenclature[F]{$F_{SV}$}{Sampling rate of Sampled Values (SV) packets from the merging unit (\si{\hertz}).}
\nomenclature[F]{$F_{GOOSE}$}{Rate of GOOSE messages from Intelligent Electronic Devices (IEDs) (\si{\hertz}).}
\nomenclature[F]{$D_{SV}$}{Downsampling factor applied to the SV data stream.}
\nomenclature[F]{$D_{GOOSE}$}{Downsampling factor applied to the GOOSE data stream.}
\nomenclature[F]{$F_{s}$}{Effective rate at the model input, defined as $\max(F_{GOOSE}\!\times\!D_{GOOSE},\, F_{SV}\!\times\!D_{SV})$.}
\nomenclature[F]{$BS$}{Batch size, i.e., the number of samples processed together by the model.}
\nomenclature[F]{$T_{a}$}{Time duration of the communications processor algorithm within the sending IED.}
\nomenclature[F]{$T_{b}$}{Transit time of the message across the substation network.}
\nomenclature[F]{$T_{c}$}{Time duration of the communications processor algorithm within the receiving IED.}
\nomenclature[F]{$T_{transmission}$}{Total message transmission time ($T_{a} + T_{b} + T_{c}$), constrained by IEC\,61850 performance classes.}
\nomenclature[F]{$T_{pre}$}{Preprocessing time (e.g., parsing and downsampling).}
\nomenclature[F]{$T_{batch}$}{Batch accumulation time, $T_{batch} = \tfrac{BS}{F_{s}}$.}
\nomenclature[F]{$T_{average}$}{Average per-sample inference time of the model on a given platform.}
\nomenclature[F]{$T_{TF}$}{Time to process one full batch of size $BS$; $T_{TF} = BS\,\times\,T_{average}$.}
\nomenclature[F]{$T_{pro}$}{Dominant processing latency for a single batch, $T_{pro} = \max(T_{batch},\, T_{TF})$.}
\nomenclature[F]{$T_{total}$}{Overall anomaly‐detection latency; $T_{total} = T_{pre} \,+\, T_{pro}$.}
\nomenclature[F]{$H$}{Hardware and framework used during inference.}

\section{Preliminaries}
\label{sec:Preliminaries}
\subsection{Transformer Architecture and In-context Learning (ICL)}
\label{subsec:ICL}

Transformers are a class of deep neural networks that have achieved state-of-the-art performance in diverse sequence modeling tasks \cite{vaswani2017attention}. Central to this architecture is the \emph{self-attention mechanism}, which efficiently captures long-range dependencies within sequential data. By stacking multiple layers of multi-head self-attention and feed-forward networks, transformers enable robust feature extraction and flexible representation learning.

Our approach adopts a \emph{decoder-only transformer} (GPT-2 \cite{radford2019language}) but specialized for network traffic analysis. Three key design elements underlie its effectiveness in sequential pattern recognition:

\begin{enumerate}
    \item \textbf{Causal Attention Mechanism.} Each prediction depends only on past observations, naturally aligning with the temporal ordering of packet streams in digital substations.
    \item \textbf{Parameter-Sharing Through Transformer Blocks.} This facilitates handling variable-length input sequences without architectural changes, crucial for evolving network traffic patterns.
    \item \textbf{Streamlined Decoder-Only Inference.} By omitting an encoder stage, the model can efficiently process real-time data while preserving strong pattern-recognition capabilities.
\end{enumerate}

Although these architectural choices draw from GPT-2’s language modeling advances, they generalize to intrusion detection by capturing temporal dependencies in packet streams. The causal attention mechanism suits event-sequential data, while the parameter sharing and decoder-only configuration reduce latency—both critical for real-time anomaly detection in digital substations.

\textbf{In-context learning (ICL).} A particularly appealing property of transformer models, including GPT-2–style decoders, is their ability to adapt to new tasks or patterns given as input context (e.g., prompt examples) \cite{brown2020language}. Formally, consider a transformer model \(M_\theta\) that processes a sequence of length \(N\). The first \(N-1\) tokens or samples, \(\{(x_1, y_1), (x_2, y_2), \ldots, (x_{N-1}, y_{N-1})\}\), serve as the \emph{in-context data} \(D_{N-1}\). The query point \(x^q\) appears in the \(N^\text{th}\) position, and the model predicts its label \(y^q\) as
\[
    y^q = M_\theta\bigl(x^q; D_{N-1}\bigr).
\]
In effect, the model produces
\[
    P(y^q \mid x^q, D_{N-1}),
\]
conditioned on the in-context data. This mechanism enables the transformer to infer novel patterns or anomalies from the prompt alone, obviating the need for retraining when new types of attacks emerge. By leveraging these architectural and learning properties, our GPT-2–based solution can efficiently detect zero-day anomalies in digital substations. 

\subsection{IEC-61850 Digital Substations}
\label{subsec:DigitalSubstations}
In the digital substations that use the IEC-61850 protocol, there are three main network protocols:  Sampled Values (SV), Generic Object-Oriented Substation Events (GOOSE), and MMS \cite{quincozes2023ereno}. In this paper, we only consider the SV and GOOSE, as they are involved in substation protection functions. SV packets transmit sampled voltage and current values from MUs to IEDs, while GOOSE packets enable fast and reliable communication between IEDs for exchanging control commands. Despite the benefits offered by the IEC-61850, it also introduces vulnerabilities that can be exploited by attackers. For example, the lack of authentication and encryption in GOOSE and SV protocols can allow attackers to inject false data, manipulate control commands, or launch denial-of-service attacks \cite{reda2021vulnerability}. These vulnerabilities highlight the need for developing effective IDS for IEC-61850-based digital substations.
As each IED continuously receives the packets, a time window of incoming packet data can be used as the in-context data for the transformer architecture (same as $D_{N-1}$ in the last section). The goal is to infer whether the next observed packet data will be an attack or not. 

\begin{figure}[!ht]
    \centering
    \includegraphics[width=\columnwidth]{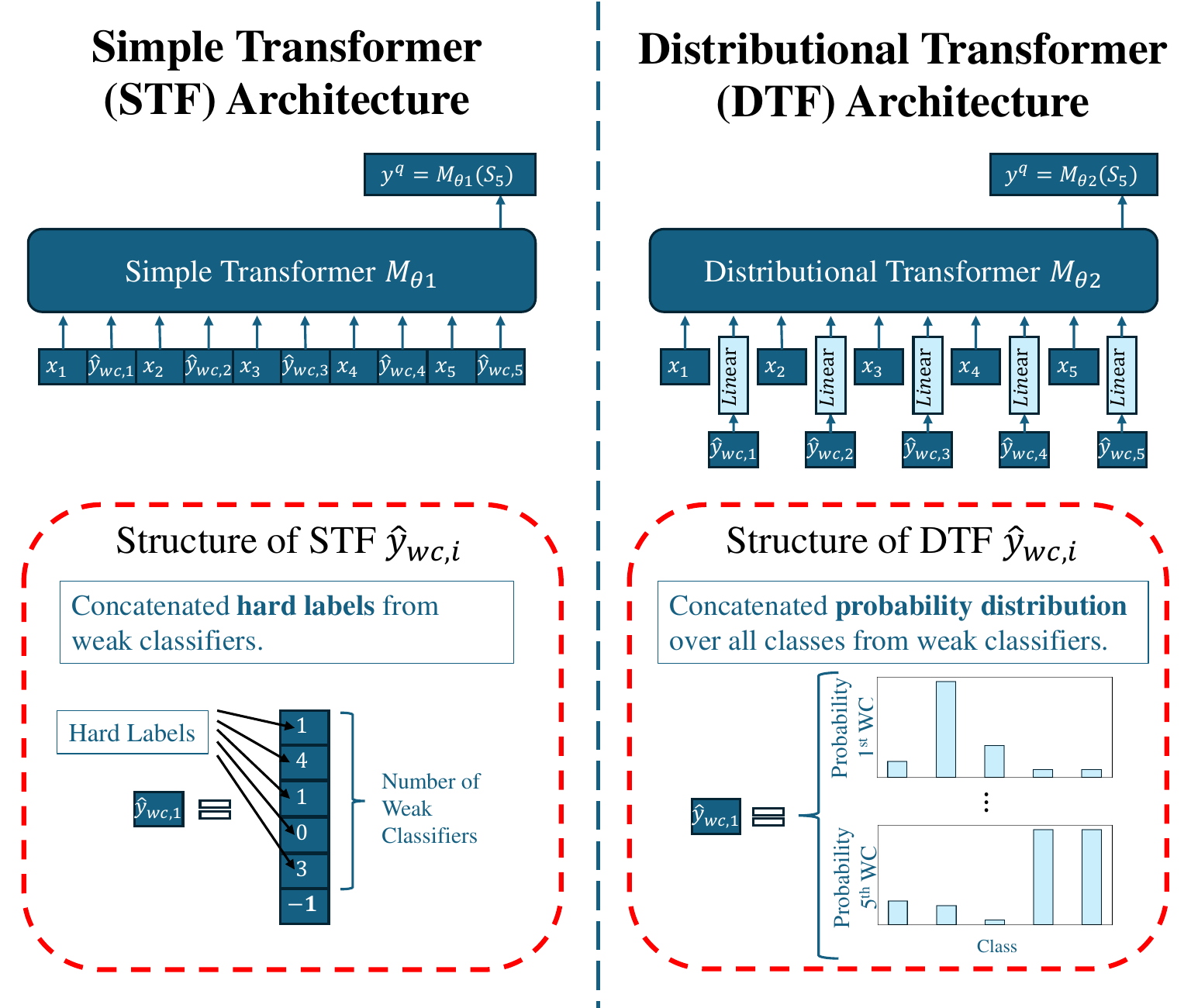}
    \vspace{-0.2cm}
    \caption{Comparison of Distributional Transformer (DTF) and Simple Transformer (STF) Architectures: DTF uses probability distributions over all classes as weak labels, while STF uses hard labels as weak labels for input.}
    \label{fig:DTFvsTF}
            \vspace{-0.4cm}

\end{figure}
\end{appendices}

\end{document}